\documentstyle[aclap]{article}
\input psfig

\begin{document}
\author{Fernando Gomez \\
Dept.~of Computer Science \\ Univ. of Central Florida \\
Orlando, FL 32816, USA \\{\tt gomez@cs.ucf.edu}\And
Richard Hull \\
Dept.~of Computer
Science \\  Univ. of Central Florida \\ Orlando, FL 32816, USA \\
{\tt hull@cs.ucf.edu}\And
Carlos Segami \\
Math. and Comp. Sci. Dept. \\ Barry
University \\ Miami Shores, FL 33161, USA\\
{\tt segami@buvax.barry.edu}}
\title{Acquiring Knowledge From Encyclopedic Texts\thanks{This 
research is being funded by NASA-KSC Contract NAG-10-0120}}
\maketitle

%%%%%%%%%%%%%%%%
%%  Abstract  %%
%%%%%%%%%%%%%%%%
\begin{abstract}
A computational model for the acquisition of knowledge from encyclopedic texts 
is described. The model has been implemented in a program, called SNOWY, that 
reads unedited texts from {\em The World Book Encyclopedia}, and acquires new 
concepts and conceptual relations about topics dealing with the dietary habits 
of animals, their classifications and habitats.  The program is also able to 
answer an ample set of questions about the knowledge that it has acquired.  
This paper describes the essential components of this model, namely semantic 
interpretation, inferences and representation, and ends with an evaluation of 
the performance of the program, a sample of the questions that it is able to 
answer, and its relation to other programs of similar nature.
\end{abstract}

%%%%%%%%%%%%%%%
%%  Section  %%
%%%%%%%%%%%%%%%
\section {Introduction}

We present an approach to the acquisition of knowledge from encyclopedic texts.
The goal of this research is to build a knowledge base about a given 
topic by reading an encyclopedic article.  Expert systems could use this 
database to tap in for pieces of knowledge, or a user could directly query 
the database for specific answers. Then, two possible applications could be 
derived from our research: (a) the automatic construction of databases from 
encyclopedic texts for problem-solvers and (b) querying an encyclopedia in 
natural language.  The idea is to build a database from an encyclopedic text
on the fly. Then, if a user asks the question, say, {\em Which bears eat 
seals?} the system would reply by saying something like ``I don't know. But, 
wait a minute, I am going to read this article and let you know." In the 
process of reading the article, the system builds a small knowledge base 
about bears and calls the question-answering system to answer any question
posed by an expert system or a human.  The long-term goal of our research is 
to read an entire article on, say bears, and to build a knowledge base about 
bears. Since this goal requires dealing with an extraordinary number of 
research issues, we have concentrated on a series of topics about animals, 
including diet, habitat, and classification.  A skimmer scans the article for 
sentences relevant to a given topic and passes these sentences to the 
understanding system, called SNOWY, for complete parsing, interpretation, 
concept formation, concept recognition and integration in long-term memory 
(LTM). But, if, in the process of learning about the dietary habits of, say 
beetles, the program is told that they cultivate fungi, and the program is 
able to interpret that sentence, that knowledge will also be integrated in 
LTM. Consequently, our approach to understanding expository texts is a 
bottom-up approach in which final knowledge representation structures
are built from the logical form of the sentences, without intervening scriptal 
or frame knowledge about the topic. Hence, our system does not start with a 
frame containing the main slots to be filled for a topic, say ``diet," as in 
recent MUC projects \cite{Sundheim:A}, but rather it will build 
everything relevant to diet from the output of the interpretation phase. Then, 
when we are talking about the topic, it will not make any difference if the 
sentences refer to what the animals eat, or what eats them. Every aspect 
dealing with the general idea of {\em ingest} can be analyzed and properly 
integrated into memory.  Our corpora for testing our ideas has been {\em The 
World Book Encyclopedia} \cite{WB:A}, which is one or two levels less complex 
than the {\em Collier's Encyclopedia}, which, in turn, is less complex than
the {\em Encyclopaedia Britannica}. 

%%%%%%%%%%%%%%% 
%%  Section  %% 
%%%%%%%%%%%%%%% 
\section{Interpretation}

In order for the integration component to integrate a concept in LTM, a 
successful parse and interpretation needs to be produced for a sentence or at 
least for one of its clauses.  The input to the interpretation phase is built 
by a top-down, lexical-driven parser\cite{GO:WUP}, which parses the sentences 
directly into syntactic cases.  Prepositional phrases are left unattached 
inside the structure built by the parser.  It is up to the interpreter to 
attach them, identify their meaning and the thematic roles that they may 
stand for. The parser is a deterministic machine, containing mechanisms for
minimizing the need for backing up.  The average amount of time in parsing a 
sentence from the encyclopedia is about one second. The parser has presently 
a lexicon of about 35,000 words.  The rate of success of producing a full 
and correct parse of a sentence is, as of this writing, 76.8\% for this 
Encyclopedia (see below for a detailed discussion of test results and machine 
used).  The parser begins the parsing of a sentence on a syntactic basis 
until the meaning of the verb is recognized.  The rules that determine the 
meaning of the verbs, called VM rules, are classified as {\em subj-rules, 
verb-rules, obj-rules, io-rules, pred-rules, prep-rules, and 
end-of-clause-rules}.  These rules are activated when the verb, a syntactic 
case or a prepositional phrase has been parsed, or when the end of the clause 
has been reached, respectively.  In most cases, the antecedents of these rules 
contain selectional restrictions which determine whether the 
interpretation of the syntactic constituent is a subclass of some concept in 
SNOWY's LTM ontology.  If during an examination of LTM the selectional 
restriction is passed, the consequent(s) of the VM rule establish the proper 
meaning or {\em verbal concept} for the verb.  If no rules fire, the 
parser inserts the syntactic case or prepositional phrase in the structure 
being built and continues parsing.  These rules, of which there are just over 
300, incorporate a considerable degree of ambiguity 
procrastination \cite{Rich:A}. For instance, rather than writing an 
{\em obj-rule} for determining the meaning of ``take" saying: {\em If LTM(obj) 
is-a medicament then meaning-of take is ingest}, which will immediately jump 
to determine the meaning of ``take" in {\em Peter took an aspirin} when the 
{\em obj} is parsed, it is better to write that rule as an {\em end-of-clause} 
rule.  This will avoid making the wrong assumption about the meaning of the 
verb in {\em Peter took an aspirin to Mary}.  

\begin{figure} 
\leavevmode\psfig{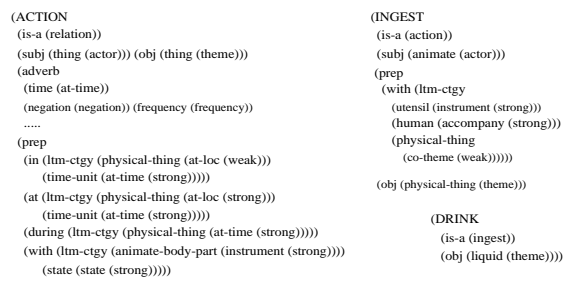}
\caption{Organization of the Verbal Concepts} 
\label{ingest-hier}
\end{figure}

\begin{figure} 
\leavevmode\psfig{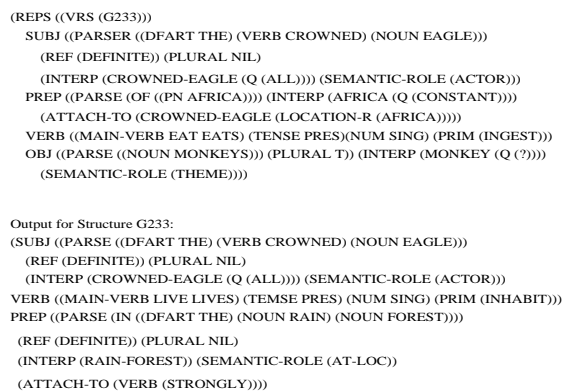}
\caption{Output of the parser and interpreter for the sentence 
 {\em The crowned eagle of Africa lives in the rain forests and eats monkeys}.}
\label{eagle-sent}
\end{figure}

When the verbal concept has been identified, the syntactic cases and
prepositional phrases already identified by the parser and any subsequent 
constituents are interpreted by matching them against the representation of 
the verbal concept.  Verbal concepts are not reduced to a small set of 
primitives; on the contrary they are organized into a classification hierarchy 
containing the most general actions near the root node and the most specific 
at the bottom. Figure 1 contains simplified examples of the root node 
{\em action}, the subconcept {\em ingest}, and one subconcept of {\em ingest}, 
{\em drink}.  The entry {\em subj} in the node {\em ingest} means that the 
{\em actor} of {\em ingest} is an animate, and that this is indicated 
syntactically by the case {\em subj}. The entries for the preposition ``with" 
mean that if the object of the PP is a utensil, then the case expressed by the
preposition is the {\em instrument} case; if the object of the preposition is 
a human, then the case is the {\em accompany} case, etc.  The entries 
{\em strong} and {\em weak} indicate whether the verbal concept claims that 
preposition strongly or weakly, respectively. This information is used by the 
interpreter to attach PPs.  The {\em drink} node has only one entry for the 
case {\em theme}.  The others are inherited from the nodes, {\em ingest} and
{\em action}.  If the context to be understood requires a more detailed 
understanding of how animals drink, distinct from how humans drink, then the 
concept {\em drink} may be split into the concepts {\em human-drink} and 
{\em animal-drink}.  Consequently, the specification of the hierarchy depends 
on the domain knowledge to be acquired.  The algorithm that matches syntactic 
cases or prepositional phrases to the representation of the verbal concept 
searches the hierarchy in a bottom-up fashion. The search ends with success or
failure, if the LTM entry in the verbal concept is true or false,
respectively.  
Hence, entries in subconcepts override the same entries in superconcepts. The
node {\em action} provides an excellent way to handle situations not
contemplated in subconcepts by defaulting them to those in the {\em
action} root node.

Figure 2 depicts the output of the interpreter. The slot subject
contains the output of the parser marked by the slot PARSE, the
interpretation of the noun phrase marked by INTERP, and the thematic
role marked by SEMANTIC-ROLE.  The entry Q within the INTERP slot
indicates the quantifier for that case.  The entries for the slot
PREP, contain, in addition to the PARSE slot, the slot ATTACH-TO
indicating which concept in the parse structure to attach that
constituent to, and the meaning of the preposition indicated, in this
case, by the subslot LOCATION-R. The quantifier for ``monkey" is a
question mark, because its value is not indicated in the text. 
In \cite{GO:SH}, the reader may find a detailed discussion of the 
interpretation issues presented in this session.

%%%%%%%%%%%%%%% 
%%  Section  %% 
%%%%%%%%%%%%%%% 
\section{Inferences} 
If the knowledge about dietary habits of animals were indicated in the
texts by using ``eat" and its cognates, the task of acquiring this
knowledge would be rather simple. But, the fact is that an
encyclopedia article may refer to the dietary habits in a variety of
manners.  Figure 3 contains a hierarchy about the diet topic. The
verbs that trigger these verbal concepts are indicated by writing
$[verb]$. The verb ``dig out" triggers the action {\em dig-r}. Those
verbal concepts from which an ingest relation is inferred are
indicated by writing an asterisk by its side.  The representation of
dig-r is:

\vspace{2mm}

\leavevmode\psfig{figure=dig.ps,height=0.8in,width=1.5in}
\vspace{3mm}

The representation of {\em dig-r} is similar to the previous verbal
concepts, except for the entry that says {\em addition-rule}.  This is
an inference rule saying that if the {\em actor} of this action is an
animal and the {\em theme} is an animate, then add to LTM the relation
saying that the {\em actor} ingests the {\em theme}.  This rule
permits the acquisition of the fact that bears eat squirrels and mice
from the sentence {\em A grizzly has long, curved claws that it uses
chiefly to dig out ground squirrels and mice}.  Similar inference
rules are stored in the other actions marked with an asterisk. Then,
if the system reads the sentence {\em Bears are fond of honey}, it will infer
that they eat honey.  The inference rules are also inherited from the
nodes representing the actions in the hierarchy.  Hence, if the verbal
concept {\em animal-fish}, shown in Figure 3, was suggested for the
sentence {\em Owls have been known to fish in shallow creeks}, we
would inherit an addition rule from the verbal concept {\em
animal-hunt} which would then infer an ingest relation.

\begin{figure}[t] 
\leavevmode\psfig{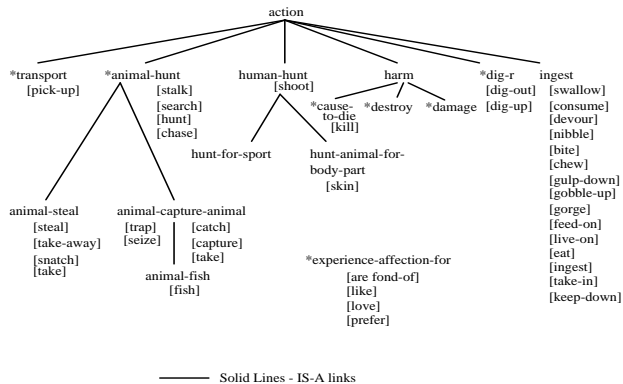}
\caption{Conceptual Verbs Organizing the Topic} 
\label{diet-hier}
\end{figure}

The hierarchy is also helpful in avoiding incorrect inferences.
Sentences discussing humans hunting animals do not automatically imply
ingest relations, especially when an explicit purpose is given which
is not ingest-related.  For example in {\em People hunt some kinds of
seals for their soft fur}, it is unlikely that the people mentioned
will eat those seals.  Therefore, we have separate verbal concepts for
hunt relations where humans are the actors, whose inference rules do
not suggest ingest relations, except in cases like {\em Eskimos hunt
polar bears for food}.  Furthermore, addition rules typically have
constraints within their antecedents to prevent inappropriate
inferences, i.e., we would not want to infer an ingest relation when
processing the sentence {\em Tigers search for warm places to sleep
during the day}, and a constraint on the theme of ``search for" to be
at least {\em animate} rejects the inference.

%%%%%%%%%%%%%%% 
%%  Section  %% 
%%%%%%%%%%%%%%% 
\section{Interpreting Noun Phrases and \protect \\ Restrictive Modifiers} 

Detecting classification relations in the text becomes a must, not
only if questions of the type {\em Which owls eat fish?}, or {\em
Which eagles eat hyraxes?} are to be answered, but also for the
acquisition of the knowledge in sentences like {\em The prey of polar
bears consists of seals}, or {\em The diet of bears consists of nuts,
berries and small rodents}. In order to achieve this, complex noun
groups and restrictive modifiers are represented by the noun group
interpreter as classification hierarchies. One of the senses of
``diet" is represented as: {\em X1 (cf (is-a (food) R1))}, where R1 is
the relation {\em ingest} with {\em actor = animal}, and {\em theme =
food}.  Then, ``the diet of bears" is represented as the concept {\em
X2 (cf (is-a (food) R2))}, where R2 is the relation {\em ingest} with
{\em actor = bear}, and {\em theme = food}. The slot {\em cf} contains
the necessary and sufficient conditions that define the concept X2.
Then, the meaning of X2 is:

\begin{center}
 $ \forall(x) (X2(x) \Longleftrightarrow Food(x) \; \; \wedge \;
R2(x))$
\end{center}

The same interpretation is given to restrictive relative clauses.  The
phrase ``eagles that live in the rain forests" is represented as
{\em X3 (cf (is-a (eagle) R3))}, where R3 is the relation ``live in the rain
forests." Once these structures are built by the interpreter, a
classifier that is a component of the integration algorithm integrates
these concepts in the proper position in LTM.  The interpretation of
complex nouns proceeds by attempting to determine the meaning of pairs
of items in the complex noun, utilizing a scheme that combines the
items in the complex noun from left to right. For example, in the
interpretation of ``big red wine bottle" an attempt is made to find a
meaning for the terms ``big red," ``big wine," ``big bottle," ``red
wine," ``red bottle" and ``wine bottle." If one item in the complex
noun can be paired (i.e., a meaning can be found) with more than one
other item in the complex noun, then the algorithm returns more than one
interpretation for the complex noun, and disambiguation routines are
activated. In our example, a meaning is found for ``big bottle," ``red
wine," ``red bottle," and ``wine bottle," from which the algorithm
returns the two possible interpretations:

\vspace{2mm}

\leavevmode\psfig{figure=bottle.ps,height=0.2in,width=2.0in}
\vspace{3mm}

Finding the meaning of terms of the form ``item1 item2" reduces to
finding a relation that connects the concepts corresponding to
``item1" and ``item2" in LTM. Consequently, this algorithm as well as
the algorithm that finds the meaning of PPs and syntactic cases
depends on a {a priori} set of concepts that constitutes the basic
ontology of SNOWY.  As SNOWY reads, it adds new concepts to this
ontology as explained below. Its initial ontology consists of 1243 concepts.

In order to find 
a semantic relation between two pair of items in a NP, the items
or any of their
superconcepts must belong to the {\em a priori} ontology.  If the
noun group interpreter does not find a semantic relation between two
items, the algorithm will hyphenate them. This has been the case for
``rain forest,'' and ``crowned eagle'' (see Figure 2). ``Rain-forest'' is
constructed in LTM as a subconcept of ``forest.'' But, no semantic
relation will be built between ``rain'' and ``forest.''  However, if the
pair of items is ``sea mammal,'' the algorithm builds {\em X3
(cf(is-a(mammal) live-in(sea)))}, because ``mammal'' and ``sea'' are
categorized in LTM as subconcepts of ``animate'' and ``habitat,''
respectively. The algorithm will produce the same representation for 
``sea mammal'' and ``mammals that live in the sea,'' except for the names of 
the concepts, which are dummy names with no meaning. The {\em recognizer} 
algorithm is able to tell that the two concepts are the same concept by 
examining the content of the {\em cf} slot, and activating a {\em classifier} 
that analyzes the subsumption relations between a pair of concepts.
In \cite{GO:EXP}, the reader may find a detailed discussion of the 
{\em recognizer} algorithm. Note that the algorithm will produce the concept 
{\em X4 (cf(is-a(lion) live-in(sea)))} if the noun group ``sea lion" is not in 
quotation marks or capitalized.

%%%%%%%%%%%%%%%
%%  Section  %%
%%%%%%%%%%%%%%%
\section{Final Knowledge Representation Structures}

The input of the interpreter is passed to the formation phase that builds the 
final knowledge representation structures. These are in turn integrated
into LTM upon activating a {\em recognizer} algorithm and an {\em integration}
algorithm both of which make extensive use of a {\em classifier} similar
to the one reported in \cite{Brach:B}.  The construction of the final 
knowledge representation structures is done as follows.  The interpretation 
phase, if successful, has built a relation, and a set of thematic roles for 
each sentence. Let us call the thematic roles of the relation the entities for 
that relation.  All the {\em n} entities of a {\em n-ary} relation are 
represented as objects in our language, and links are created pointing to the 
representation of the relation, which is represented as a separate structure, 
called an {\em a-structure}.  Figure 4 depicts the representation produced 
from the interpretation of the sentence {\em The crowned eagle of Africa lives 
in the rain forests and eats monkeys}.  Five objects have been created; 
CROWNED-EAGLE, AFRICA, RAIN-FOREST, MONKEY and @X235, which stands for the 
concept ``crowned eagle of Africa." The relation @A237 represents the 
{\em ingest} relation between the object @X235 and the object MONKEY.  The 
object MONKEY points to this structure by the entry under  MONKEY that says
{\em ingest\%by (@x235 (\$more (@a237))))}, and the object @X235 also points to 
this structure by the slot {\em (ingest (monkey (\$more (@a237))))}.
The scope of the quantifiers is from left to right. Then, the meaning of 
structure @A239 (assuming that the question mark in the quantifier slot of 
MONKEY stands for ``some" as the question-answering assumes) is $\forall(x) 
(@X235(x) \Longrightarrow \exists y (MONKEY(y) \; \wedge \; INGEST(x,y)))$. 
An {\em a-structure} can be linked to other {\em a-structures} by slots 
expressing {\em causality}, {\em time}, etc., as becomes necessary in the 
representation of {\em Birds migrate south when it freezes}.

\begin{figure}
\leavevmode\psfig{figure=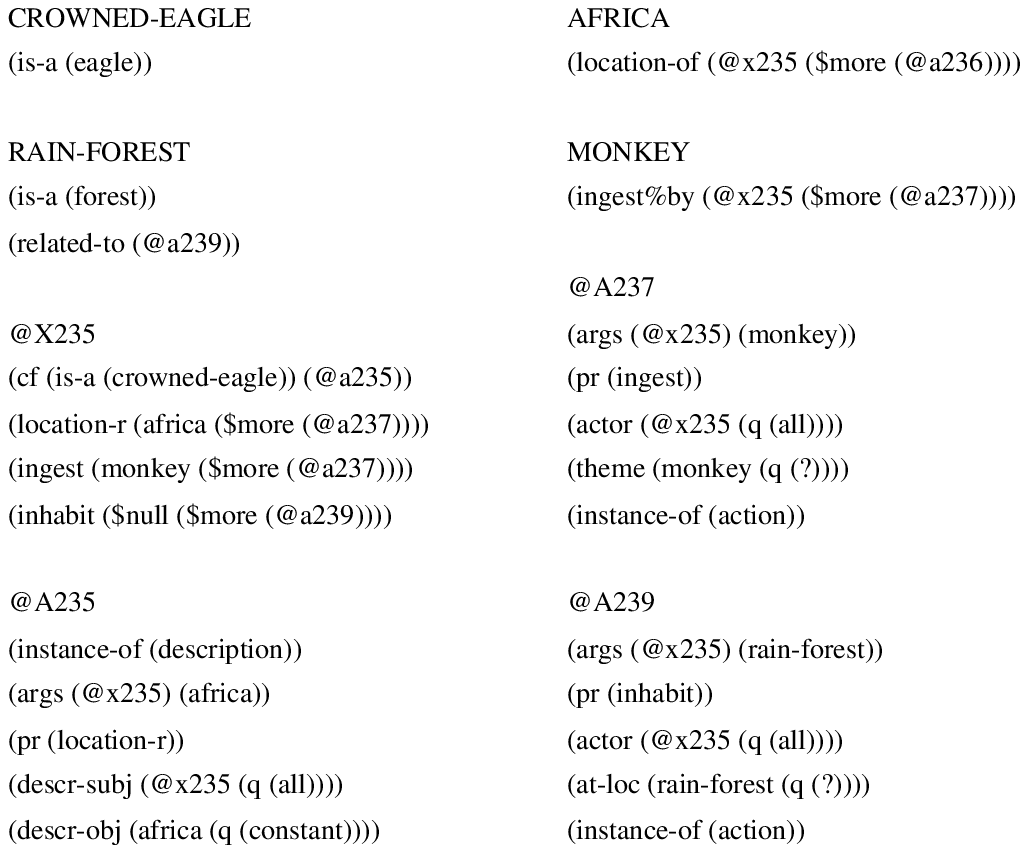,height=2.0in,width=3.0in}
\caption{Formation Structures}
\label{eagle-form}
\end{figure}

%%%%%%%%%%%%%%%
%%  Section  %%
%%%%%%%%%%%%%%%
\section{Results}

Table 1 below provides statistics revealing how well the system performed
during testing.  The system was initially trained on ten articles: bears,
beavers, beetles, elephants, frogs, penguins, raccoons, seals, snakes, and 
tigers.  Then, two more articles about sharks and eagles were analyzed to 
assess our progress.  Test texts were chosen randomly by a student selecting 
a letter of the alphabet and then finding texts about animals within those 
volumes of the encyclopedia.  The letter ``B" and the letter ``M" were chosen.
The texts were then selected from those volumes.  In December of 1993, an 
article about birds was chosen. No component of the system (lexicon, parser, 
interpreter, etc.) was pre-prepared with information about this article. 
The lexicon of the parser consisted of 10,000 words.  This text was the 
largest article that the system had analyzed, containing approximately 1330 
sentences and 16,000 words. None of the designers of the system read this 
article prior to the test. And even if they had read it, it would have been 
of very little use because the system has reached such complexity that it
is not easy to assess how it is going to perform in an article of 1330 
sentences.

The first row of Table 1 indicates that 145 sentences were selected from the 
bird text by a keyword/pattern-based skimmer.  Of these 145 sentences, 91 were 
relevant to the dietary habits domain. A total of 23 relevant sentences were 
missed, primarily due to keywords or patterns not contemplated.  An example of 
sentence that is relevant but was not selected is {\em Robins and sparrows, 
for example, are highly effective against cabbageworms, tomato worms, and leaf 
beetles}.  The parser was able to produce a correct parse for 58 of the 91 
relevant sentences (64\%), even in cases where the sentence contained unknown 
words.  Among the sentences successfully parsed, the parser encountered some 
40 unknown words of which approximately 70\% were names of birds, such as 
``grosbeaks", ``flycatchers", ``titmice", ``thrashers," etc., 
and 30\% were common words.  Of those 58 parsed sentences, 32 (55\%) were 
interpreted correctly.  The output for these 32 sentences was then passed 
to the formation, recognition, and integration phases to be inserted into LTM.
Interpretation failures can be attributed to missing VM rules, comparatives, 
and problems of anaphoric reference.

Later, in April of 1994, three more texts were randomly chosen for testing.
A summary of the results for those three tests is also given in Table 1.
In this test, the parser ran with a lexicon consisting of about 35,000 words.  
The improvement of the interpreter was mainly due to new interpretation rules, 
additions to the hierarchy of verbal concepts, and to the hierarchy of 
concepts that organize the inferences about the topic.

All of the tests were run on a SPARC Classic Machine executing Allegro Common 
Lisp.  The average time to completely process a selected sentence on this 
platform was 3.1 seconds.  This is a conservative figure because it includes 
the processing time of the skimmer, i.e., some amount of overhead is necessary 
for file handling and for determining when a sentence is irrelevant. 
Therefore, the average time for processing a sentence is actually less than 
3.1 seconds.

\begin{table} 
\caption{Statistics for the Sentences of the Bird, Bat, Monkey, and Mouse Texts}
\vspace{5 mm}
\leavevmode\psfig{figure=result-table.ps,height=1.5in,width=3.0in}
\end{table}

The following is a list of natural language questions posed to the system after
reading the bird, bat, and monkey articles and the contents of the 
system-generated answers. Note that the system output has been altered and
condensed for the sake of brevity.  In answering the questions, the system 
uses classification-based reasoning, not theorem proving. Many complex chains 
of inferences can be obtained by keeping memory organized in a principled 
manner.  In those cases in which a question asks about a concept that does not
exist in LTM, the {\em classifier} is activated to place that concept in LTM 
and obtain an answer.  See \cite{GO:SMC} for a detailed discussion of all 
these issues and the theorems proving the soundness of the inference 
algorithms.

\noindent
{\em What do birds eat?} sapsucker ingest tree-sap; hummingbird ingest nectar;
duck ingest plant-matter, grass, seaweed; louisiana-water-thrush ingest 
water-insect; young-bird ingest earthworm, insect, small animal; ........

\noindent
{\em Which birds eat nectar?}
  hummingbird ingest nectar 

\noindent
{\em What kinds of insect eaters are there?}
  chickadee, creeper, flycatcher, kinglet, swallow, swift, thrasher,  
  titmice, vireo, warbler, woodpecker, owl 

\noindent
{\em What is gravel?}
  I don't know, but I know that: 
    bird ingest gravel $<$ related-to $>$ bird *assist* grinding-process

\noindent
{\em Do most cactus dwellers eat insects?}
  yes

\noindent
{\em What kills birds?}
  eagle is-a bird, and hunters and trappers kill eagles;
  osprey is-a bird, and hunters and trappers kill osprey

\noindent
{\em When do most birds search for food?}
  at-time day

\noindent
{\em Do birds help people?}
  yes, bird help farmer

\noindent
{\em How do birds help farmers?}
  bird ingest $<$insect which ingest crop$>$; bird ingest weed-seed 

\noindent
{\em Do bats eat blood?}
  yes, some bat eat blood because
    vampire-bat is-a bat and
    vampire-bat ingest blood

\noindent
{\em How much blood do vampire bats eat?}
  vampire-bat ingest blood quantity 1 tablespoon *frequency* day

\noindent
{\em Do vampire bats attack human beings?}
  yes, vampire bat harm human *frequency* sometimes

\noindent
{\em Do monkeys have enemies?}
  yes, some monkey has-enemy cheetah hyena jackal leopard lion
  because
    $<$monkeys inhabit at-loc ground$>$ has-enemy cheetah hyena jackal leopard lion

%%%%%%%%%%%%%%%
%%  Section  %%
%%%%%%%%%%%%%%%
\section{Related Work}
In \cite{Sond:A}, frame-like structures, KL-ONE structures in fact, are also 
used to guide semantic interpretation in an application domain.  However,
the overall approach to the interpretation task presented here differs from 
that work. In approaching the problem of unrestricted texts, we agree with 
those researchers \cite {Hobbs:A,Grish:A} who think that it is possible to 
build correct parses and interpretations for real-world texts. In fact, it 
is hard for us to see how statistical methods \cite{deM:A,Church:A} could be 
used for building knowledge-bases with sufficient expressive power to 
correctly answer questions posed by expert systems or human users. We think 
that the same critique applies to skimmers \cite{Lehnert:A}, but for very 
different reasons. In order to guarantee the correctness of the knowledge-base 
built, every element in the sentence needs to be interpreted. For instance, 
if the adverb ``mostly" is not interpreted in the sentence {\em These owls eat 
mostly rodents}, the integrity of the knowledge-base built is not going to 
suffer greatly. But, if we are talking about the adverb ``rarely" in the 
sentence {\em These owls rarely eat rodents}, the situation becomes much more 
serious, as we found out.

This work has advanced a new approach to semantic interpretation that occupies 
a middle ground between those approaches that rely heavily on the parser for 
building structures and attaching PPs, subordinate clauses, etc. 
\cite{Grish:A,Hobbs:B,Tomita:A} and semantic-centered approaches
\cite{Riesbeck:A,Cardie:A,Slator:A}.  Our parser delegates all the 
burden of dealing with structural ambiguity (attachment of PPs, relative 
clauses, subordinate clauses, etc.) to the interpreter. That is one of the 
reasons why it is so fast. The interpreter has a very sophisticated algorithm 
that uses the information built in the verbal concepts in order to attach PPs. 
Yet, if the parser does not build a parse, albeit a shallow one, the 
interpreter will not know what to do. Moreover, the interpreter does not 
question the parser when it says this constituent is an {\em obj}, or 
{\em subj}, or a {\em time-np}, etc. This is a situation that we are not happy 
about, because the parser identifies some constituents incorrectly, especially 
the {\em time-np}. We are studying mechanisms under which the interpreter 
will {\em override} the parser and will get it out of trouble in processing 
very complex sentences \cite{Krupka:A,Jacobs:A}.

%%%%%%%%%%%%%%%
%%  Section  %%
%%%%%%%%%%%%%%%
\section{Conclusions}
We have presented a method for the acquisition of knowledge from encyclopedic 
texts. The method depends on understanding what is being read, which in turn 
depends on: (1) providing a successful parse and interpretation for a sentence,
(2) building final knowledge representation structures from the logical form 
of the sentence, which involves creating new concepts and relations as the 
system is reading, and (3) integrating in LTM those concepts and relations 
that the recognizer algorithm fails to recognize, which in many cases involves
the reorganization of concepts in LTM.

The results that are reported in this paper are very encouraging, because
a high percentage of the failures are due to some incomplete implementations
of some aspects of the system.  For instance, in dealing with anaphora we 
have incorporated in our system some of the ideas reported in 
\cite{Grosz:A,Hobbs:B}, but our work is clearly insufficient in that regard.
A major hole in the interpreter, as of this writing, is that it
does not interpret comparatives,
except very simple ones like quantifiers, e.g., ``more than 2.'' The
interpreter needs to have mechanisms in place to recover the elliptical
elements in comparatives, which in many cases require solving
extrasentential reference, e.g., 
{\em The golden eagle defends a territory of about 20 to 60 square
miles. The bald eagle holds a smaller territory}.
The skimmer uses very rudimentary techniques, and there is a lot of room for 
improvement here. In any case, this has not been a major concern of this
research. Moreover, because the system is so incredibly fast, if the skimmer
overgenerates, it is not much of a problem.
An aspect related to the skimmer that we have no space to discuss is
that it became necessary to build an algorithm to recognize 
subclasses of the class of animals being searched for every
question. For instance, if the 
question is {\em Do sharks eat plankton}, this algorithm analyzes {\em every} 
sentence in the encyclopedic article, before being passed to the skimmer, 
searching for NPs denoting subclasses of sharks.  This is necessary because 
the author may introduce the concept ``Mako sharks" in a context unrelated 
to the relation {\em ingest}, and consequently, the skimmer is not going to 
select this sentence. Then, if the author later on says {\em Makos feed
on other
fish, including herring, mackerel, and swordfish}, the system has no way to 
relate ``Mako" to sharks, missing the fact that sharks feed on herring, mackerel,
and swordfish.  In June, we tested the parser on 2900 sentences from 25 
articles, including non-animal articles such as blackholes,
cancer, computer chips, napoleon, greenhouse effect, etc., and the rate of 
success was 76.8\%. However, some sentences
are still not parsed because some subcategorizations of verbs are wrong
or incomplete, or the phrasal lexicon is incomplete. We are
confident that the parser may reach a plateau at 85\% or 90\% for
this Encyclopedia.
The remaining
10\% or 15\% may require considerable help from the interpreter to be parsed.

%%%%%%%%%%%%%%%%%%%%
%%  Bibliography  %%
%%%%%%%%%%%%%%%%%%%%

\end{document}